\documentclass{article}

\PassOptionsToPackage{numbers, sort}{natbib}

\usepackage[preprint]{neurips_2026}

\usepackage[utf8]{inputenc} 
\usepackage[T1]{fontenc}    
\usepackage{hyperref}       
\usepackage{url}            
\usepackage{booktabs}       
\usepackage{amsfonts}       
\usepackage{nicefrac}       
\usepackage{microtype}      
\usepackage{xcolor}         
\usepackage{amsmath}
\usepackage{amssymb}
\usepackage{pifont}    
\usepackage{array}    
\newcommand{\cmark}{\ding{51}} 
\newcommand{\xmark}{\ding{55}} 
\usepackage[table]{xcolor} 
\newcommand{\up}[1]{$_{+#1}$}
\usepackage{graphicx}
\usepackage{cleveref} 
\usepackage{booktabs, multirow}
\usepackage{makecell} 
\usepackage{colortbl}
\usepackage{xcolor}
\usepackage{float}

\crefname{appendix}{Appendix}{Appendices}

\title{Learn where to Click from Yourself: \\On-Policy Self-Distillation for GUI Grounding}

\author{%
  Yan Zhang$^{1,3,}$\thanks{Equal contribution} \quad 
  Daiqing Wu$^{1,3,}$\footnotemark[1] \quad 
  Huawen Shen$^{1,3}$ \quad 
  Can Ma$^{1,3,}$\thanks{Corresponding authors}  \quad 
  Yu Zhou$^{2,}$ \footnotemark[2] \\ 
  $^1$Institute of Information Engineering, Chinese Academy of Sciences \\
  $^2$VCIP \& TMCC \& DISSec, College of Computer Science, Nankai University \\
  $^3$School of Cyber Security, University of Chinese Academy of Sciences \\
  \vspace{0.15cm}
  \texttt{zhangyan2022@iie.ac.cn; yzhou@nankai.edu.cn} \\
}

\begin{document}

\maketitle

\begin{abstract}

Graphical User Interface (GUI) grounding maps natural language instructions to the visual coordinates of target elements and serves as a core capability for autonomous GUI agents. Recent reinforcement learning methods (e.g., GRPO) have achieved strong performance, but they rely on expensive multiple rollouts and suffer from sparse signals on hard samples. These limitations make on-policy self-distillation (OPSD), which provides dense token-level supervision from a single rollout, a promising alternative. However, its applicability to GUI grounding remains unexplored. In this paper, we present GUI-SD, the first OPSD framework tailored for GUI grounding. First, it constructs a visually enriched privileged context for the teacher using a target bounding box and a Gaussian soft mask, providing informative guidance without leaking exact coordinates. Second, it employs entropy-guided distillation, which adaptively weights tokens based on digit significance and teacher confidence, concentrating optimization on the most impactful and reliable positions. Extensive experiments on six representative GUI grounding benchmarks show that GUI-SD consistently outperforms GRPO-based methods and naive OPSD in both accuracy and training efficiency. Code and training data are available at \url{https://zhangyan-ucas.github.io/GUI-SD/}.


\end{abstract}

\section{Introduction}
\begin{figure}[t]
\begin{center}
\includegraphics[width=1.0\textwidth]{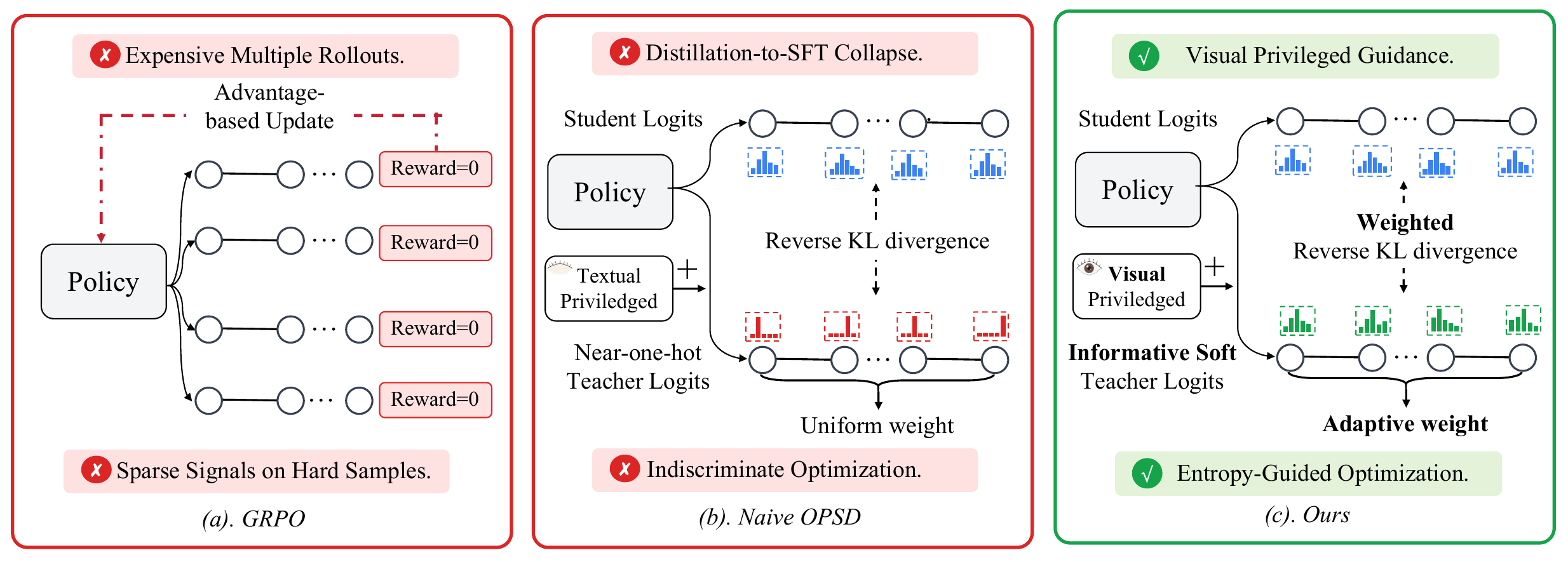}
\end{center}
\caption{(a) \textbf{GRPO} requires expensive multiple rollouts and produces zero reward on hard samples. (b) \textbf{Naive OPSD} forwards the policy twice and distills via reverse KL between student and teacher logits with uniform per-token weight $w=1.0$, yet suffers from distillation-to-SFT collapse and indiscriminate optimization. (c) \textbf{Ours} addresses both issues via visual privileged guidance and entropy-guided optimization.}

\label{fig:figure1}

\end{figure}

Autonomous GUI agents have emerged as a promising direction for human-computer interaction, where GUI grounding serves as the fundamental capability of mapping natural language instructions to visual coordinates of target elements~\cite{gou2024navigating,cheng2024seeclick}. To this end, a growing body of work~\cite{hyperclick,gui-eyes,Gui-actor,gui-bee} has adopted reinforcement learning for GUI grounding, among which GRPO-based methods \cite{gui-g1,infigui-r1} have become the dominant paradigm as shown in \Cref{fig:figure1}(a). Specifically, given a user instruction, GRPO \cite{guo2025deepseek,shao2024deepseekmath} encourages the policy model to explore diverse solutions by sampling multiple rollouts, and evaluates each with a designed verifiable reward, such as binary \cite{infigui-r1}, distance-constrained \cite{se-gui}, or gaussian-based feedback \cite{GUI-G$^2$}. The advantage of each rollout is then computed relative to the group reward distribution, steering the policy to reinforce successful explorations while discouraging unsuccessful ones \cite{yuan2025enhancing}. 

Despite the advances, GRPO-based training for GUI grounding still depends on expensive multiple rollouts to estimate advantages and suffers from sparse signals on hard samples where all rollouts receive zero reward. These limitations call for a paradigm that can deliver dense supervision from fewer interactions. Recently emerging on-policy self-distillation (OPSD) \cite{yang2026self, shenfeld2026self,qu2026pope,hubotter2026reinforcement,song2026expanding} offers such a possibility, providing token-level supervision from a single rollout by deploying the same model as both teacher and student under asymmetric contexts. Specifically, the asymmetry lies in the privileged information that is accessible to the teacher but hidden from the student, such as reference solutions \cite{shenfeld2026self}, verifier signals \cite{qu2026pope}, and environment feedback \cite{hubotter2026reinforcement}. Guided by this privileged context, the teacher acts as a stronger model, yielding a more reliable output distribution whose per-token log-probabilities form a reverse Kullback-Leibler (KL) divergence loss that continuously refines the student. By replacing sparse outcome-level rewards with dense token-level guidance, OPSD provides an appealing alternative for improving both training efficiency and supervision quality.


Motivated by these advantages, we explore for the first time the application of OPSD to GUI grounding. 
However, as illustrated in \Cref{fig:figure1}(b), directly adapting OPSD to this setting encounters two critical bottlenecks: 1) \textbf{Distillation-to-SFT Collapse.} Naive OPSD paradigm directly appends the target coordinate as text to the teacher's input, causing the teacher's supervisory distribution to collapse into near-one-hot targets with near-zero entropy. In this regime, minimizing the KL divergence between teacher and student becomes equivalent to minimizing cross-entropy against hard labels, effectively reducing distillation to supervised fine-tuning (SFT) and erasing the dark knowledge \cite{hinton2015distilling} that makes soft-label supervision beneficial. 2) \textbf{Indiscriminate Optimization.} Naive OPSD applies reverse KL to distill all tokens uniformly, yet higher-order coordinate digits steer the optimization direction far more effectively than lower-order digits. Furthermore, the teacher's confidence varies across tokens, and treating all tokens equally propagates unreliable signals from low-confidence positions, leading to sub-optimal gradients.

To address these issues, we propose GUI-SD (\textbf{GUI} Grounding via \textbf{S}elf-\textbf{D}istillation), an OPSD framework tailored for GUI grounding, which combines visually enriched privileged context with an entropy-guided loss to deliver rich token-level supervision for precise coordinate generation. Specifically, GUI-SD builds the teacher's privileged context by highlighting the ground-truth region with a bounding box and applying a Gaussian soft mask that gradually fades the surrounding areas. Paired with an instructional hint, this visual prompt delivers informative yet constrained prior knowledge, guiding the teacher to the target without leaking the exact coordinates. Furthermore, GUI-SD introduces entropy-guided distillation, an adaptive objective that replaces uniform token weighting with targeted supervision. It prioritizes higher-order coordinate digits that dominate grounding accuracy while amplifying supervision from confident teacher predictions.

Extensive experiments across six representative grounding benchmarks (ScreenSpot-v2 \cite{atlas}, ScreenSpot-Pro \cite{screenspotpro}, UI-Vision \cite{uivision}, MMBench GUI L2 \cite{mmbench}, OSWorld-G\cite{osworldg}, and OSWorld-G-Refine \cite{osworldg}) demonstrate that GUI-SD substantially outperforms GRPO-based methods \cite{infigui-r1,se-gui,GUI-G$^2$} and naive OPSD \cite{shenfeld2026self} in both accuracy and training efficiency. Detailed ablation studies further validate that visually enriched privileged context provides effective teacher guidance, while entropy-guided distillation concentrates optimization on the most impactful coordinate tokens.

Our main contributions are summarized as follows:
\begin{itemize}
    \item To the best of our knowledge, we present the first exploration of the OPSD framework in the GUI grounding domain, offering an appealing alternative to GRPO-based methods that suffer from expensive multiple rollouts and sparse signals on hard samples.
    \item We propose GUI-SD, which integrates visually grounded teacher guidance with entropy-aware distillation, enabling rich and reliable supervision that concentrates optimization on the most impactful coordinate tokens.
    \item Extensive experiments verify the effectiveness of GUI-SD over naive OPSD and GRPO-based methods across six representative GUI grounding benchmarks, demonstrating significant improvements in both accuracy and training efficiency, establishing OPSD as a promising paradigm for future GUI grounding research.
\end{itemize}

\section{Preliminary}
\textbf{OPD.} While GRPO-type reinforcement learning has driven significant progress in GUI grounding, its sparse sequence-level rewards provide no dense token-level guidance, offer little or even zero feedback on difficult samples, and require heavy online sampling \cite{ui-inst}. On-Policy Distillation (OPD) \cite{song2026survey} offers an alternative paradigm, where a separate, typically larger, teacher model $\pi_{\hat{\theta}}$ provides token-level supervision along the student's sampled trajectories. By distilling the teacher's output distribution at each decoding step, OPD delivers continuous learning signals that enable more sample-efficient training and meaningful gradient updates even for samples that would otherwise receive no reward.

\textbf{OPSD.} To remove the dependence on a separate teacher, On-Policy Self-Distillation (OPSD) \cite{zhang2026opsdl} deploys the same model $\pi_\theta$ as both teacher and student, with the two roles operating under asymmetric contexts. Specifically, the teacher is granted access to privileged information $r$ (e.g., ground-truth answers \cite{shenfeld2026self} or verified reasoning traces \cite{qu2026pope}) that is unavailable to the student, yielding more informative token-level distributions along the student's sampled trajectories. Formally, given the sample $(x, r)$, where $x$ denotes the input query and $ r$ the privileged context, the student generates an on-policy trajectory under $x$. Meanwhile, the teacher, conditioned on both $(x, r)$, produces step-wise target distributions along the same trajectory. Training then minimizes the per-token divergence between the student and teacher distributions at each decoding step:

\begin{equation}
\label{eq:1}
\begin{gathered}
\begin{aligned}
  \text{Student:} \quad & P_S(y_{t}) \triangleq \pi_\theta(y_{t} \mid x,\, y_{<t}), \\
  \text{Teacher:} \quad & P_T(y_{t}) \triangleq \pi_\theta(y_{t} \mid x,\, r,\, y_{<t}),
\end{aligned} \\[6pt]
\mathcal{L}(\theta) = \mathbb{E}_{y\sim P_S}\!\left[\frac{1}{|y|}\sum_{t=1}^{|y|} D_{\mathrm{KL}}\!\big(P_S(y_{t})\ \big\|\ P_T(y_{t})\big)\right],
\end{gathered}
\end{equation}
where $D_{\mathrm{KL}}$ measures the KL divergence, $y_{<t}$ denotes the generated trajectory up to step $t$, $y_t$ denotes the token generated at step $t$, and $|y|$ is the total length of the trajectory.

\section{Empirical Analysis of OPSD for GUI Grounding}
\label{sec:empirical}
To understand the failure of naive OPSD in GUI grounding, we analyze its teacher supervision signal through two complementary entropy-based views. Following prior distillation studies \cite{wang2026learning, hinton2015distilling, kim2026trust,jung2025todi, su2025ea}, we consider: (1) sample-level entropy, which assesses whether the overall teacher distribution remains informative or collapses toward near-one-hot targets \cite{wang2026learning, hinton2015distilling, kim2026trust}; and (2) token-level entropy, which evaluates the reliability of supervision across coordinate digits \cite{jung2025todi, su2025ea}.


\subsection{Distillation-to-SFT Collapse at Sample-level}
Prior OPSD methods provide the teacher with privileged information in textual form, such as reference solutions, verifier signals, or environment feedback. Following this design, a naive adaptation to GUI grounding feeds the ground-truth coordinate directly into the teacher input as text. We evaluate the resulting teacher signal at the sample level using two statistics over coordinate digits: average entropy and average top-1 probability. As shown in Table~\ref{tab:signal_comparison}, the naive OPSD teacher produces a nearly deterministic distribution, with an average entropy of only 0.17 and an average top-1 probability of 0.82. Prior distillation studies~\cite{hinton2015distilling, kim2026trust} suggest that such low-entropy targets behave similarly to hard labels, causing divergence minimization to degenerate toward standard cross-entropy training. This interpretation is consistent with downstream performance: naive OPSD improves over SFT by only 1.5\% on ScreenSpot-Pro (55.6 vs. 54.1), indicating that textual privilege yields little benefit beyond hard-label supervision. In contrast, the optimized privileged context in GUI-SD produces a more informative teacher signal, reaching 60.7\% on ScreenSpot-Pro, a 6.6\% improvement over SFT, demonstrating that a well-designed privilege context can unlock substantial gains beyond hard-label supervision.

\textbf{Finding 1:} In GUI grounding, textual privileged information drives the teacher distribution toward near-zero entropy, largely collapsing distillation into near-SFT behavior. We term this failure mode Distillation-to-SFT Collapse.

\begin{table}[t]
\centering
\caption{Comparison between naive OPSD teacher supervisory signal and SFT signal. SSP Acc denotes model accuracy on ScreenSpot-Pro.}
\label{tab:signal_comparison}
\resizebox{0.7\textwidth}{!}{
\begin{tabular}{lccc}
\toprule
& \textbf{Avg Entropy} & \textbf{Avg Top-1 Prob} & \textbf{SSP Acc} \\
\midrule
SFT Signal                  & 0.00 & 1.00 & 54.1 \\
Teacher Signal (Naive OPSD) & 0.17 & 0.82 & 55.6\\
\textcolor{gray}{Teacher Signal (GUI-SD)} & \textcolor{gray}{0.50} & \textcolor{gray}{0.59} & \textcolor{gray}{60.7} \\
\bottomrule
\end{tabular}}
\end{table}

\subsection{Indiscriminate Optimization at Token-level}
\begin{figure}[]
\begin{center}
\includegraphics[width=1.0\textwidth]{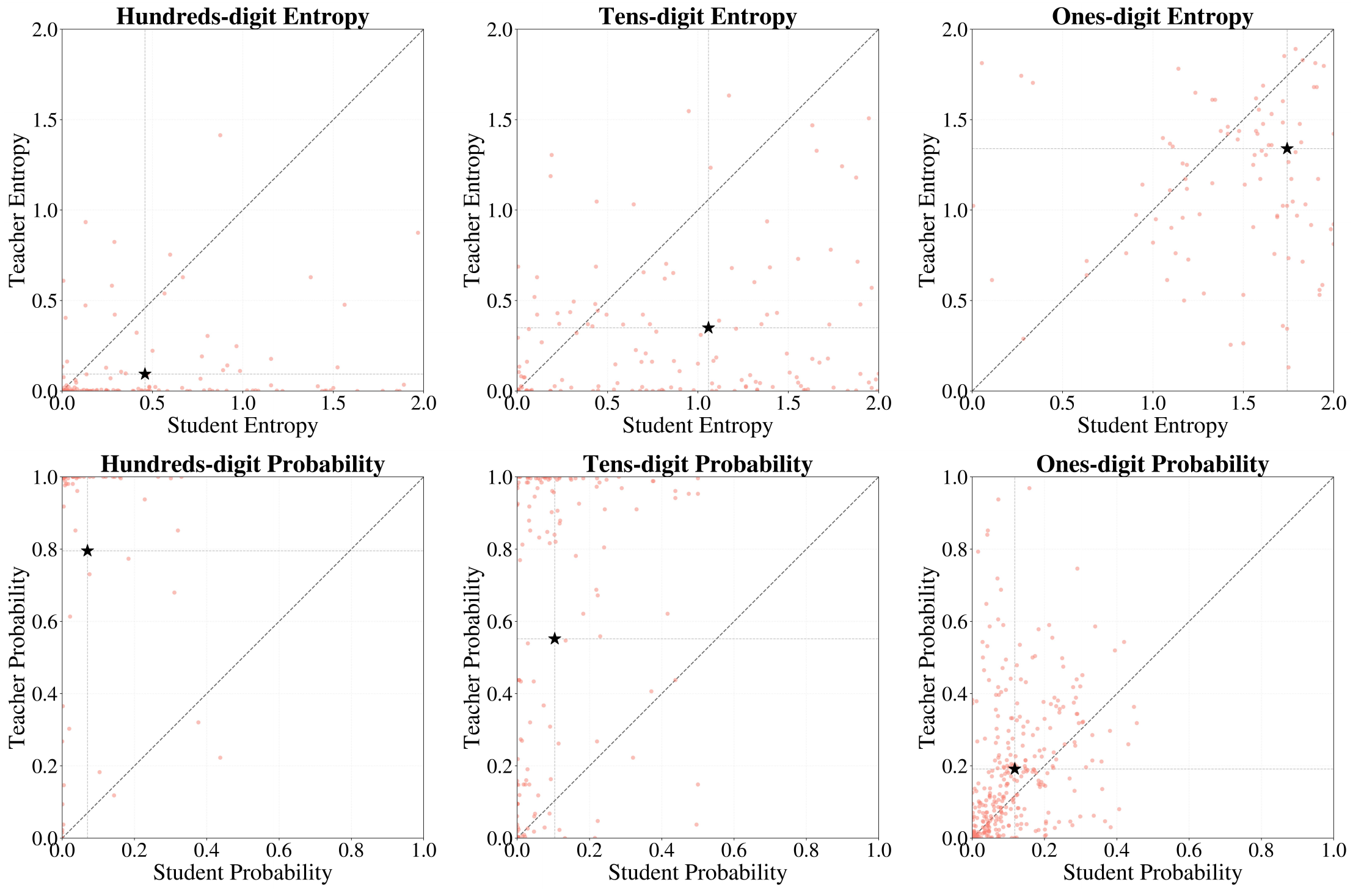}
\end{center}
\vspace{-10pt} 
\caption{Per-token analysis of teacher and student predictions on incorrectly predicted tokens across digit positions (hundreds, tens, units). Top: teacher vs. student prediction entropy. Bottom: teacher vs. student ground-truth probability. Stars denote the mean teacher and student values for each digit position.}
\label{fig:empirical}
\vspace{-10pt} 
\end{figure}


A second question concerns the token level: under a well-designed privileged context, is the teacher's supervision equally reliable across coordinate digits? To investigate this, we analyze incorrectly predicted tokens at each digit position, where teacher guidance is most needed.

As shown in \Cref{fig:empirical} (Top), the teacher generally exhibits lower entropy than the student at each digit position, indicating that its token-level preferences are sharper and thus more likely to be amplified during distillation, even on positions where the student predicts incorrectly. Notably, entropy increases from hundreds to tens to units for both teacher and student, suggesting that supervision becomes progressively less certain on lower-order digits. This pattern is echoed in \Cref{fig:empirical} (Bottom), where the teacher assigns higher probability to the ground-truth token than the student at every digit position, with the largest gap observed at the hundreds digit.

\textbf{Finding 2:} Teacher supervision in GUI grounding is inherently position-dependent: higher-order digits carry stronger and more reliable signals than lower-order digits. A uniform reverse-KL objective ignores this heterogeneity and therefore not only misallocates optimization budgets, but also risks amplifying erroneous token preferences on more uncertain digit positions. We term this failure mode Indiscriminate Optimization.


\section{Method}
\begin{figure}[t]
\begin{center}
\includegraphics[width=1.0\textwidth]{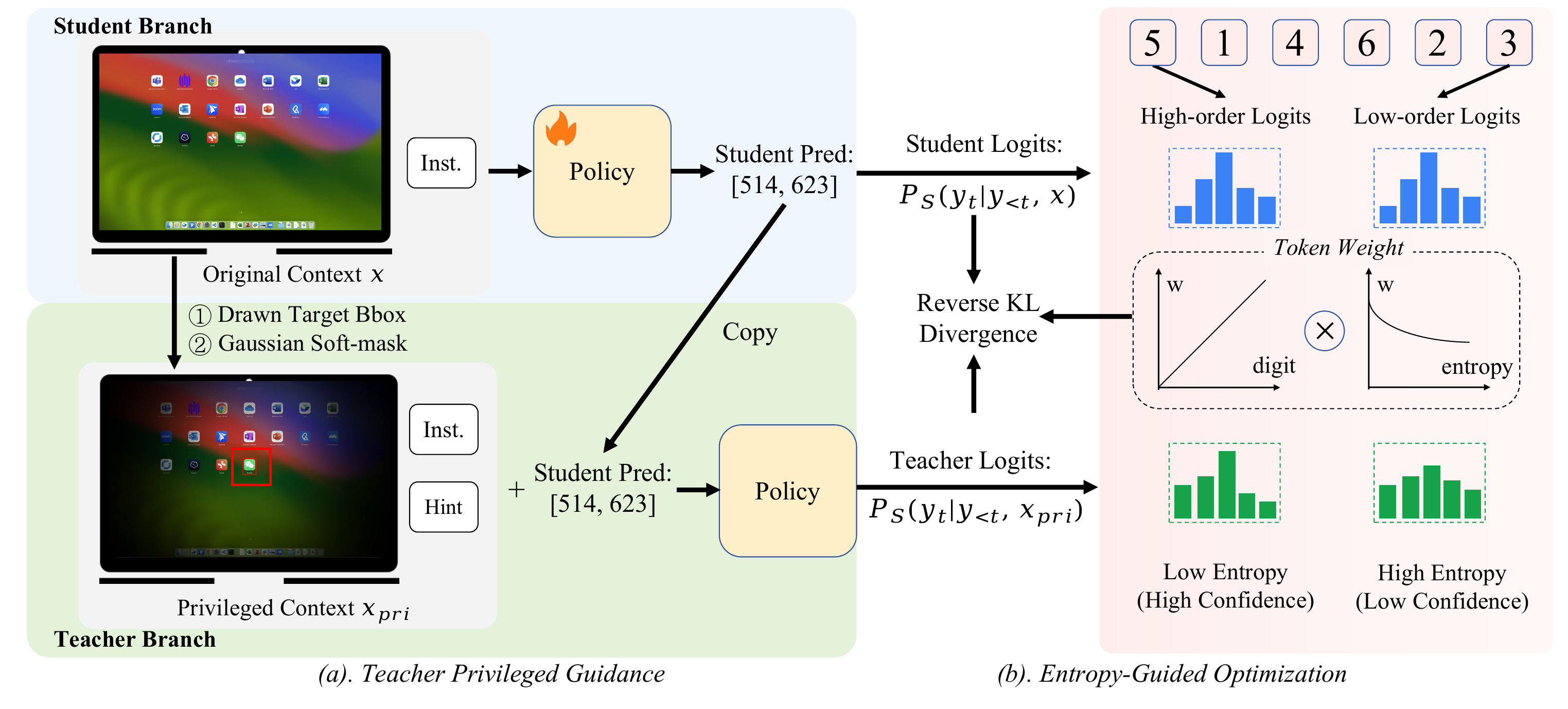}
\end{center}
\vspace{-5pt} 
\caption{Overview of the GUI-SD framework. (a) The teacher branch receives a privileged context $x_{pri}$, which augments the student's original input $x$ with a target bounding box, a Gaussian soft mask, and a hint prompt, while the student branch operates on the original context only. (b) GUI-SD trains the student with a weighted reverse-KL objective between teacher and student token distributions, where the token weight $w(t)$ prioritizes high-order tokens via positional credit assignment and filters unreliable supervision via entropy-gated supervision.
}
\label{fig:method}
\vspace{-5pt} 
\end{figure}

Building on the above analysis, we propose GUI-SD as a targeted solution to the two failure modes of naive OPSD in GUI grounding. As illustrated in \Cref{fig:method}, GUI-SD comprises two complementary components. (a) \textbf{Visual Privileged Guidance} replaces textual privilege with a visually enriched privileged context, enabling the teacher to maintain informative soft distributions rather than collapsing toward near-one-hot targets. (b) \textbf{Entropy-Guided Optimization} replaces uniform reverse KL with adaptive token weighting, so that optimization emphasizes high-value coordinate digits while downweighting uncertain supervision.


\subsection{Visual Privileged Guidance}

Finding~1 shows that textual privileged information collapses the teacher distribution toward near-one-hot targets, reducing distillation to near-SFT behavior. To address this issue, GUI-SD replaces textual coordinate privilege with visually grounded privileged context, allowing the teacher to receive target-aware guidance without direct access to the exact coordinate.

Specifically, as shown in \Cref{fig:method}(a), the teacher input is augmented with a red bounding box over the ground-truth region and a lightweight hint prompt, e.g., ``Hint: The answer is located within the red rectangle.'' This combination supplies informative yet constrained prior knowledge, guiding the teacher toward the target while preserving non-degenerate soft supervision. In addition, to improve localization in high-resolution GUI scenes with dense layouts, we apply a Gaussian soft mask that progressively attenuates image regions farther from the annotated target, effectively creating a zoom-in effect around the privileged region. The modulation factor for each pixel is defined as:
\begin{equation}
\alpha(x, y) = \exp\left(-\frac{d^2}{2\sigma^2}\right),
\end{equation}
where $d$ denotes the distance from pixel $(x, y)$ to the nearest edge of the ground-truth bounding box (with $d=0$ inside the box), and $\sigma$ is scaled according to the target size with a minimum floor to prevent over-masking small objects. This design preserves full visibility of the target region while smoothly suppressing irrelevant surrounding content.

\subsection{Entropy-Guided Optimization}
In GUI-SD, the privileged information $r$ in the general OPSD formulation (\cref{eq:1}) is instantiated as the visually enriched context $x_{pri}$. Conditioned on $x_{pri}$, the teacher produces step-wise target distributions to provide supervisory signals along the student trajectory.

Finding~2 shows that uniform per-token distillation is suboptimal for GUI grounding: coordinate digits differ substantially in both positional importance and supervision reliability. GUI-SD therefore replaces the uniform objective with an entropy-guided weighted reverse-KL loss:
\begin{equation}
\mathcal{L}(\theta) = \mathbb{E}_{y\sim P_S}\!\left[\frac{1}{|y|}\sum_{t=1}^{|y|} w(t) \cdot D_{\mathrm{KL}}\!\big(P_S(y_{t})\ \big\|\ P_T(y_{t})\big)\right],
\end{equation}
where
\begin{equation}
w(t) = w^{\text{pos}}(t) \cdot w^{\text{ent}}(t).
\end{equation}




\paragraph{Positional Credit Assignment.}
The first factor, $w^{\text{pos}}(t)$, captures the positional significance of coordinate digits. In decimal coordinate prediction, an error in a higher-order digit induces a much larger spatial deviation than an error in a lower-order digit. For example, an incorrect hundreds digit may introduce roughly $100$ pixels of error, whereas an incorrect units digit affects only about $1$ pixel. To reflect this positional asymmetry, we assign decaying weights from the most significant digit exponentially to the least significant digit:
\begin{equation}
w^{\text{pos}}(t) = \alpha \cdot k_t,
\end{equation}
where $t$ is the token index in the generated sequence, $k_t \in \{1, 2, 3, 4\}$ denotes the digit position of token $t$ (1 for units, 2 for tens, 3 for hundreds, 4 for thousands), and $\alpha > 0$ is a scaling factor. For non-numeric tokens, we set $w^{\text{pos}}(t)=1$.

\paragraph{Entropy-Gated Supervision.}
The second factor, $w^{\text{ent}}(t)$, captures the reliability of teacher supervision. Since the teacher's confidence varies across tokens, uniformly distilling all positions may over-amplify uncertain teacher preferences and propagate unreliable gradients. We therefore use the entropy of the teacher distribution to modulate supervision strength:
\begin{equation}
w^{\text{ent}}(t) = \exp\!\left( -\frac{H\!\left(p_T(\cdot \mid x_{pri}, y_{<t})\right)}{\tau} \right),
\end{equation}
where $H(\cdot)$ denotes the per-token entropy of the teacher distribution and $\tau$ is a scaling factor. Tokens with low entropy receive stronger supervision, whereas uncertain predictions are automatically down-weighted.

\begin{table}[htbp]
\centering
\renewcommand{\arraystretch}{1.2} 
\caption{GUI grounding accuracy on six benchmarks including ScreenSpot-V2 (SS2), ScreenSpot-Pro (SSP), MMBenchGUI (MMG), UI-Vision (UIV), OSWorld-G (OSW-G), and OSWorld-G\_R (OSW-GR). Bold indicates the best results. Training time is measured in hours per epoch on 8$\times$A800-80G. The detailed experimental results on each benchmark are in the \Cref{app:detailed_results}.}
\resizebox{\textwidth}{!}{%
\begin{tabular}{l cccccccc}
\toprule
\textbf{Method} & \textbf{Time/epoch} & \textbf{SSP} & \textbf{SS2} & \textbf{UIV} & \textbf{OSW-G} & \textbf{OSW-GR} & \textbf{MMG} & \textbf{Avg.} \\ 
\midrule
UI-TARS-7B \cite{uitars}      & -    & 35.7 & 91.6 & 17.6 & -    & -    & -    & -    \\
GTA1-7B \cite{gta1}          & -    & 50.1 & 92.4 & -    & 55.1  & 67.7    & -    & -    \\
GUI-Actor-7B \cite{Gui-actor}         & -    & 44.6 & 89.5 & -    & -    & -    & -    & -    \\
TongUI-7B \cite{tongui}       & -    & 24.7 & 88.7 & 18.0 & -    & -    & -    & -    \\
JEDI-7B  \cite{osworldg}         & -    & 39.5 & 91.7 & 25.2 & -    & -    & -    & -    \\
InfiGUI-R1-7B \cite{infigui-r1}    & -    & 35.7 & -    & -    & -    & -    & -    & -    \\
SE-GUI-7B   \cite{se-gui}      & -    & 47.3 & 90.3 & -    & -    & -    & -    & -    \\
LPO-7B    \cite{LPO}        & -    & -    & 90.5 & -    & -    & -    & -    & -    \\
GUI-G$^2$-7B  \cite{GUI-G$^2$}       & -    & 47.5 & 93.3 & -    & -    & -    & -    & -    \\
HyperClick-7B \cite{hyperclick}    & -    & 48.2 & 93.7 & 25.7 & -    & -    & 79.6 & -    \\
UI-Ins-7B  \cite{ui-inst}     & -    & 52.2 & 94.0   & -    & -    & -    & 83.1 & -    \\
ZoomUI-7B  \cite{zoomui}       & -    & 52.8 & -    & 27.1 & 54.2 & -    & 72.8 & -    \\ 
ZwZ-8B \cite{ZwZ} & -    & 56.8 &  - & - & 60.0  & 69.0 & - & -    \\ 
MolmoWeb-Ground-8B \cite{gupta2026molmoweb} & -    & - & 91.8    & - & - & -    & - & -    \\ 
Propose-then-Critic-8B  \cite{Propose-then-Critic}    & -    & 58.7 & 91.3    & 28.9 & 59.6 & -    & 78.4 & -    \\ 
\midrule
Qwen3-VL-Instruct \cite{qwen3vl} & -    & 53.6 & 93.2 & 25.2 & 58.7 & 67.4 & 83.0 & 63.5 \\
\textit{+ GRPO-Binary} \cite{infigui-r1}   & 16.9 & 56.8 & 94.6 & 27.6 & 61.2 & 68.6 & 84.3 & 65.5\up{2.0} \\
\textit{+ GRPO-Distance} \cite{se-gui}    & 16.7 & 56.6 & 93.8 & 27.5 & 62.1 & 69.9 & 83.3 & 65.5\up{2.0} \\
\textit{+ GRPO-Gaussian} \cite{GUI-G$^2$} & 16.8 & 57.4 & 94.0 & 28.2 & 61.9 & 70.0 & 83.7 & 65.9\up{2.4} \\
\rowcolor{gray!20} \textit{+ Ours} & \textbf{4.2} & \textbf{60.7} & \textbf{95.1} & \textbf{33.3} & \textbf{64.0} & \textbf{70.9} & \textbf{86.7} & \textbf{68.4}\up{4.9} \\ 
\bottomrule
\end{tabular}%
}
\label{tab:table2}
\end{table}

\section{Experiment}
\subsection{Experimental Setup}
We conduct experiments on top of Qwen3-VL-Instruct-8B \cite{qwen3vl}, using approximately 7K samples drawn from the ScaleCUA GUI datasets \cite{scalecua} for training. We compare GUI-SD against the following baselines. \textbf{GRPO-Binary} is a standard RLVR method \cite{infigui-r1} that uses a sparse reward, yielding 1 when the prediction falls inside the target bounding box and 0 otherwise. \textbf{GRPO-Distance} computes a dense reward based on the normalized distance between the click point and the center point of the target bounding box, following SE-GUI \cite{segui}. \textbf{GRPO-Gaussian} models GUI elements as continuous Gaussian distributions to provide dense reward signals, following GUI-G$^2$ \cite{GUI-G$^2$}. Details of the training setting of GUI-SD are provided in \Cref{app:training_details}.

We comprehensively evaluate GUI-SD on six representative GUI grounding benchmarks: ScreenSpot-v2 \cite{atlas}, ScreenSpot-Pro \cite{screenspotpro}, UI-Vision \cite{uivision}, MMBench GUI L2 \cite{mmbench}, OSWorld-G \cite{osworldg}, and OSWorld-G-Refine \cite{osworldg}. Together, these benchmarks cover diverse application scenarios, hierarchical instruction following, and cross-platform generalization across different operating systems. More details on each benchmark are provided in \Cref{app:benchmarks}.

\subsection{Main Results}
\paragraph{Comparisons with Baselines.} \Cref{tab:table2} presents the evaluation results across six representative GUI grounding benchmarks. GUI-SD achieves the highest average accuracy while demonstrating superior training efficiency compared to existing GRPO-based methods. We attribute this superiority to dense token-level credit assignment and requiring only a single rollout of the policy model instead of multiple. The former provides fine-grained supervision at every decoding step, particularly beneficial on ScreenSpot-Pro and UI-Vision, where small targets on high-resolution screenshots and diverse desktop layouts across 83 applications both demand precise spatial reasoning, achieving gains of +3.3\% and +5.1\% over GRPO-Gaussian, respectively. The latter substantially reduces training cost, achieving approximately 4× faster training per epoch compared to GRPO-based methods. 


\paragraph{Comparisons with SOTA Methods.} GUI-SD surpasses existing GUI grounding methods in average accuracy across six representative benchmarks. On ScreenSpot-Pro, GUI-SD achieves 60.7\%, outperforming Propose-then-Critic \cite{Propose-then-Critic}, which relies on test-time scaling with substantial inference overhead. On OSWorld-G-Refine, GUI-SD reaches 70.9\%, surpassing ZwZ \cite{ZwZ}, which depends on large teacher models (e.g., Qwen3-VL-235B) for distillation, requiring significantly more computational resources. Notably, GUI-SD achieves these improvements through self-distillation without relying on test-time scaling or external large-scale models, demonstrating that dense token-level supervision from a well-designed privileged context can be more effective.

\begin{table}[t]
\centering
\caption{Ablation studies on teacher privileged context. We evaluate student 
performance on ScreenSpot-Pro (SSP) under varying teacher guidance settings, along with teacher supervised signal quality: sample accuracy (Acc), sample-averaged entropy (Ent.), and sample-averaged top-1 probability (Top-1). 
$\Delta$ OPSD denotes the performance difference relative to 
Naive OPSD (Row 2). ``Orig.'' denotes Original, ``Inst.'' is 
Instruction, and ``Gauss.'' is Gaussian.}
\resizebox{0.84\textwidth}{!}{%
\renewcommand{\arraystretch}{1.4}
\begin{tabular}{ll | cc | ccc}
\toprule
\multicolumn{2}{c|}{\textbf{Teacher's Guidance Setting}} & 
\multicolumn{2}{c|}{\textbf{Student}} & 
\multicolumn{3}{c}{\textbf{Teacher Signal}} \\
\cmidrule(lr){1-2} \cmidrule(lr){3-4} \cmidrule(lr){5-7}
\textbf{Visual Context} & \textbf{Text Context} & \textbf{SSP} & \textbf{$\Delta$ OPSD} & \textbf{Acc} & \textbf{Ent.} & \textbf{Top-1} \\ 
\midrule
Orig. Image              & Inst.              & 53.0 & -2.6 & 52.0 & 0.59 & 0.53 \\
Orig. Image              & Inst. + Text BBox  & 55.6 & 0    & 93.0 & 0.17 & 0.82 \\
Orig. Image + Drawn BBox & Inst. + Drawn Hint & 59.9 & +4.3 & 89.8 & 0.53 & 0.61 \\
Gauss. Zoom + Drawn BBox & Inst. + Drawn Hint & \textbf{60.7} & \textbf{+5.1} & \textbf{99.6} & \textbf{0.50} & \textbf{0.59} \\ 
\bottomrule
\end{tabular}%
}
\label{tab:table3}
\end{table}

\begin{table}[t]
\centering
\caption{Ablation study of entropy-guided distillation components. We report ScreenSpot-Pro overall accuracy, hundreds-digit accuracy, and performance on the hard subset, which consists of samples that the base model fails to predict correctly across all 8 rollouts. Entropy-gated and Positional Credit denote entropy-gated supervision and positional credit assignment, respectively.}
\label{tab:ablation}
\resizebox{0.94\textwidth}{!}{%
\begin{tabular}{ccccc}
\toprule
\textbf{Entropy-gated }& \textbf{Positional Credit} & \textbf{Screenspot-Pro} & \textbf{Hundreds-digit Accuracy} & \textbf{Hard Subset} \\ 
\midrule
\xmark & \xmark & 58.3 & 77.0 & 17.5 \\
\xmark & \cmark & 59.6 & 78.7 & 19.2 \\
\cmark & \xmark & 59.2 & 78.0 & 19.9 \\
\cmark & \cmark & \textbf{60.7} & \textbf{79.7} & \textbf{21.1} \\ 
\bottomrule
\end{tabular}}
\vspace{-10pt}
\end{table}

\subsection{Ablation Studies}

\paragraph{Effectiveness of Teacher Visual Context.}
As shown in \Cref{tab:table3}, we evaluate student performance on ScreenSpot-Pro under varying teacher guidance settings, along with teacher signal quality (sample accuracy, entropy, and top-1 probability). In Row 1, the teacher and student receive identical inputs without any privilege, yielding the weakest student score (53.0\%), confirming that asymmetric context is critical for effective self-distillation. Row 2 (Naive OPSD) appends the ground-truth coordinate as text to the teacher's input, reaching 55.6\%. However, although teacher accuracy reaches 93.0\%, the entropy drops to merely 0.17, indicating a near-collapsed distribution that reduces soft-label distillation to hard-label SFT. To address this, Row 3 delivers ground-truth information through the visual channel --- drawing a bounding box on the teacher's input image with an instructional hint. This forces the teacher to reason over the image rather than copying the answer, raising entropy to 0.53 and producing a substantially softer distribution that enables richer gradient signals, lifting the student to 59.9\% (+4.3\%). Row 4 further introduces a Gaussian soft-mask zoom-in that suppresses surrounding content, pushing teacher accuracy to 99.6\% while maintaining healthy entropy (0.50), achieving the best student performance of 60.7\% (+5.1\% over Naive OPSD).

\paragraph{Effectiveness of Entropy-guided Optimization.} As shown in \Cref{tab:ablation}, we ablate the individual components of entropy-guided distillation and report performance on ScreenSpot-Pro, its corresponding hard subset, and hundreds-digit accuracy. The first row serves as the baseline optimized solely with the standard reverse KL loss, yielding the lowest performance across all metrics. Introducing the digit-position weighting (Row 2) directly improves the hundreds-digit accuracy from 77.0\% to 78.7\%, which consequently lifts the overall ScreenSpot-Pro score to 59.6\%, confirming that the targeted optimization of leading digits is highly effective for GUI grounding. Incorporating only the entropy-guided weighting (Row 3) steadily improves the overall score to 59.2\% and yields substantial gains on the hard subset (from 17.5\% to 19.9\%). Finally, combining both components (Row 4) yields complementary gains, achieving the best results across the board: 60.7\% on ScreenSpot-Pro, 79.7\% on hundreds-digit accuracy, and a substantial jump to 21.1\% on the hard subset.

\subsection{Training Dynamics}
\begin{figure}[t]
\begin{center}
\includegraphics[width=0.9\textwidth]{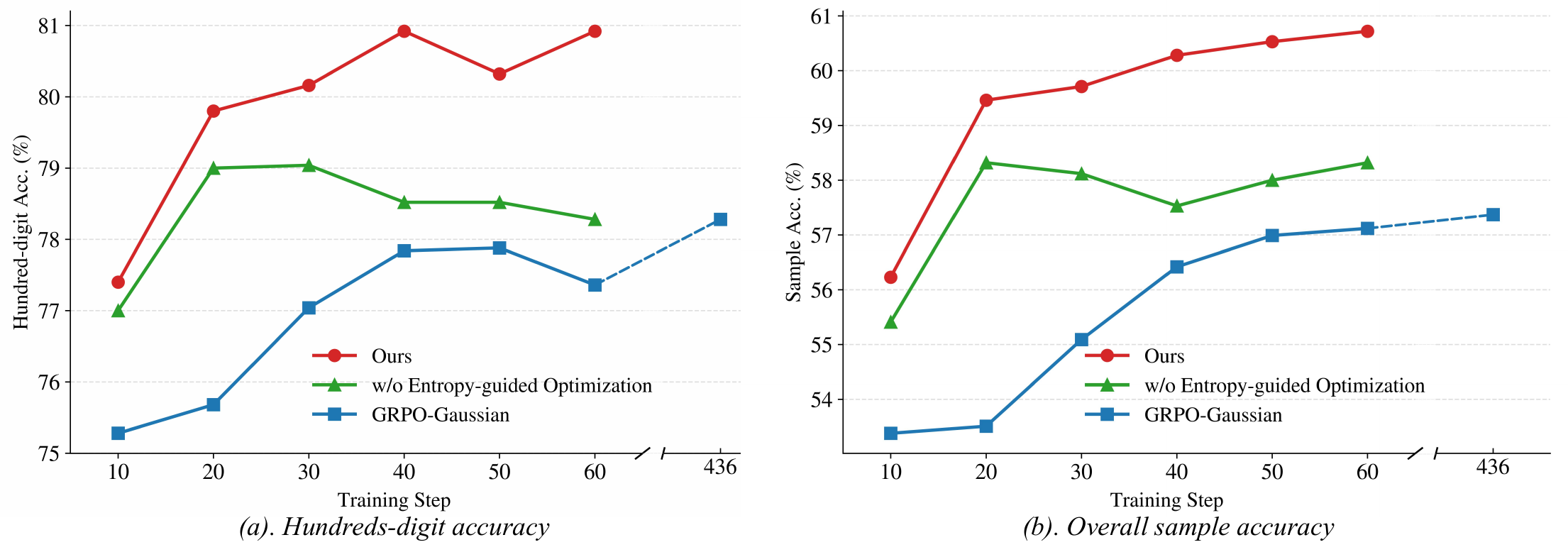}
\end{center}
\vspace{-10pt} 
\caption{Training dynamics of GUI-SD, Standard Reverse KL, and GRPO-Gaussian over optimization steps within a single epoch. 
(a) Hundreds-digit accuracy. 
(b) Overall sample accuracy.}
\label{fig:figure4}
\vspace{-10pt} 
\end{figure}

\Cref{fig:figure4} illustrates the training dynamics over the optimization steps within a single epoch, comparing GUI-SD against two baselines: GRPO-Gaussian and an ablated variant without entropy-guided optimization. As shown in \Cref{fig:figure4}(a), GUI-SD achieves significantly higher hundreds-digit accuracy than both GRPO-Gaussian and the ablated variant, which we attribute to positional credit assignment that concentrates optimization on leading digits and entropy-gated supervision that amplifies high-confidence teacher signals. As shown in \Cref{fig:figure4}(b), this improved precision on leading digits translates directly into superior overall sample accuracy. Notably, GUI-SD reaches higher performance in substantially fewer training steps than GRPO-Gaussian, validating the efficiency of dense token-level supervision over sequence-level rewards for GUI grounding.

\section{Related Work}

\subsection{On-Policy Self Distillation}
OPD has recently emerged as an effective on-policy training paradigm for delivering rich, token-level feedback \cite{song2026survey,li2026rethinking}. Specifically, the student first generates a rollout, which is subsequently fed into a stronger teacher model to provide step-by-step guidance. OPSD eliminates the need for an external teacher by having the same model serve both roles, conditioned solely on different input contexts, such as reference solutions, verifier signals, and environment feedback. A representative work, SDPO \cite{hubotter2026reinforcement}, exemplifies this by casting the feedback-conditioned model as a self-teacher, distilling its enriched next-token predictions back into the student policy. RLVR \citep{yang2026self} extends this by leveraging self-distillation to obtain token-level supervision for fine-grained update magnitudes, concurrently deriving reliable update directions from environmental feedback.  However, previous explorations of OPSD are predominantly confined to the natural language domain. When directly applied to visual tasks, especially GUI grounding, OPSD encounters critical issues such as distillation-to-SFT collapse, as detailed in Section~\ref{sec:empirical}.

\subsection{GUI Grounding via Verifiable Reinforcement Learning}
Reinforcement learning with verifiable rewards (RLVR) methods, such as GRPO, have become an effective paradigm for post-training reasoning models by enabling them to autonomously explore solution trajectories under verifiable feedback \cite{guo2025deepseek,computerrl,mobilerl}. Recent efforts have extended this paradigm to GUI grounding \cite{huang2025mobileipl,fan2026webfactory,kang2025guirlvg,zeng2026fdc,zhao2026co}. GUI-R1~\cite{gui-r1} and UI-R1~\cite{ui-r1} adopt a binary reward based on whether the predicted coordinate falls inside the target bounding box. To mitigate binary feedback limitations, GUI-G1~\cite{gui-g1} introduces dense rewards with size-based difficulty coefficients, and GUI-G$^2$~\cite{GUI-G$^2$} further models GUI elements as continuous Gaussian distributions for precise spatial alignment. Despite these improvements, GRPO-based training remains hindered by expensive multiple rollouts, sparse signals on hard samples, and heavy reliance on manually designed rewards.

\section{Conclusion and Limitations}
In this paper, we present GUI-SD, the first exploration of on-policy self-distillation for GUI grounding, which integrates visually grounded teacher guidance with entropy-guided optimization to deliver targeted token-level supervision for precise coordinate generation. Extensive evaluations across six benchmarks demonstrate that GUI-SD substantially outperforms both GRPO-based methods \cite{infigui-r1,se-gui,GUI-G$^2$} and naive OPSD \cite{shenfeld2026self,zhang2025btl,xie2025scaling} in accuracy while achieving approximately 4$\times$ faster training. One limitation is that we have not yet explored scaling to larger models or other model families beyond Qwen3-VL. A promising future direction is extending GUI-SD to long-horizon GUI agent tasks, where multi-step planning and sequential interactions introduce additional challenges.

\bibliographystyle{plainnat} 
\bibliography{references}

@article{yang2026self,
  title={Self-Distilled RLVR},
  author={Yang, Chenxu and Qin, Chuanyu and Si, Qingyi and Chen, Minghui and Gu, Naibin and Yao, Dingyu and Lin, Zheng and Wang, Weiping and Wang, Jiaqi and Duan, Nan},
  journal={arXiv preprint arXiv:2604.03128},
  year={2026}
}

@article{hinton2015distilling,
  title={Distilling the knowledge in a neural network},
  author={Hinton, Geoffrey and Vinyals, Oriol and Dean, Jeff},
  journal={arXiv preprint arXiv:1503.02531},
  year={2015}
}

@article{kim2026trust,
  title={Trust the uncertain teacher: distilling dark knowledge via calibrated uncertainty},
  author={Kim, Jeonghyun and Kim, SooKyung and Xuan, Richeng and Cho, Hyunsoo},
  journal={arXiv preprint arXiv:2602.12687},
  year={2026}
}

@article{wang2026learning,
  title={Learning While Staying Curious: Entropy-Preserving Supervised Fine-Tuning via Adaptive Self-Distillation for Large Reasoning Models},
  author={Wang, Hao and Gu, Hao and Piao, Hongming and Gong, Kaixiong and Ye, Yuxiao and Yue, Xiangyu and Han, Sirui and Guo, Yike and Wu, Dapeng},
  journal={arXiv preprint arXiv:2602.02244},
  year={2026}
}

@inproceedings{jung2025todi,
  title={Todi: Token-wise distillation via fine-grained divergence control},
  author={Jung, Seongryong and Yoon, Suwan and Kim, DongGeon and Lee, Hwanhee},
  booktitle={Proceedings of the 2025 Conference on Empirical Methods in Natural Language Processing},
  pages={8089--8102},
  year={2025}
}

@inproceedings{su2025ea,
  title={EA-KD: Entropy-based adaptive knowledge distillation},
  author={Su, Chi-Ping and Tseng, Ching-Hsun and Pu, Bin and Zhao, Lei and Yang, Jiewen and Chen, Zhuangzhuang and Lee, Shin-Jye},
  booktitle={Proceedings of the IEEE/CVF International Conference on Computer Vision},
  pages={731--740},
  year={2025}
}

@article{gui-g1,
  title={Gui-g1: Understanding r1-zero-like training for visual grounding in gui agents},
  author={Zhou, Yuqi and Dai, Sunhao and Wang, Shuai and Zhou, Kaiwen and Jia, Qinglin and Xu, Jun},
  journal={arXiv preprint arXiv:2505.15810},
  year={2025}
}

@inproceedings{GUI-G$^2$,
  title={GUI-G$^2$: Gaussian Reward Modeling for GUI Grounding},
  author={Tang, Fei and Gu, Zhangxuan and Lu, Zhengxi and Liu, Xuyang and Shen, Shuheng and Meng, Changhua and Wang, Wen and Zhang, Wenqi and Shen, Yongliang and Lu, Weiming and others},
  booktitle={Proceedings of the AAAI Conference on Artificial Intelligence},
  volume={40},
  number={39},
  pages={33214--33222},
  year={2026}
}

@article{infigui-r1,
  title={Infigui-r1: Advancing multimodal gui agents from reactive actors to deliberative reasoners},
  author={Liu, Yuhang and Li, Pengxiang and Xie, Congkai and Hu, Xavier and Han, Xiaotian and Zhang, Shengyu and Yang, Hongxia and Wu, Fei},
  journal={arXiv preprint arXiv:2504.14239},
  year={2025}
}

@article{segui,
  title={Enhancing visual grounding for gui agents via self-evolutionary reinforcement learning},
  author={Yuan, Xinbin and Zhang, Jian and Li, Kaixin and Cai, Zhuoxuan and Yao, Lujian and Chen, Jie and Wang, Enguang and Hou, Qibin and Chen, Jinwei and Jiang, Peng-Tao and others},
  journal={arXiv preprint arXiv:2505.12370},
  year={2025}
}

@inproceedings{screenspotpro,
  title={Screenspot-pro: Gui grounding for professional high-resolution computer use},
  author={Li, Kaixin and Meng, Ziyang and Lin, Hongzhan and Luo, Ziyang and Tian, Yuchen and Ma, Jing and Huang, Zhiyong and Chua, Tat-Seng},
  booktitle={Proceedings of the 33rd ACM International Conference on Multimedia},
  pages={8778--8786},
  year={2025}
}

@article{osworldg,
  title={Scaling computer-use grounding via user interface decomposition and synthesis},
  author={Xie, Tianbao and Deng, Jiaqi and Li, Xiaochuan and Yang, Junlin and Wu, Haoyuan and Chen, Jixuan and Hu, Wenjing and Wang, Xinyuan and Xu, Yuhui and Wang, Zekun and others},
  journal={arXiv preprint arXiv:2505.13227},
  year={2025}
}

@article{uivision,
  title={Ui-vision: A desktop-centric gui benchmark for visual perception and interaction},
  author={Nayak, Shravan and Jian, Xiangru and Lin, Kevin Qinghong and Rodriguez, Juan A and Kalsi, Montek and Awal, Rabiul and Chapados, Nicolas and {\"O}zsu, M Tamer and Agrawal, Aishwarya and Vazquez, David and others},
  journal={arXiv preprint arXiv:2503.15661},
  year={2025}
}

@article{mmbench,
  title={Mmbench-gui: Hierarchical multi-platform evaluation framework for gui agents},
  author={Wang, Xuehui and Wu, Zhenyu and Xie, JingJing and Ding, Zichen and Yang, Bowen and Li, Zehao and Liu, Zhaoyang and Li, Qingyun and Dong, Xuan and Chen, Zhe and others},
  journal={arXiv preprint arXiv:2507.19478},
  year={2025}
}

@article{atlas,
  title={Os-atlas: A foundation action model for generalist gui agents, 2024},
  author={Wu, Zhiyong and Wu, Zhenyu and Xu, Fangzhi and Wang, Yian and Sun, Qiushi and Jia, Chengyou and Cheng, Kanzhi and Ding, Zichen and Chen, Liheng and Liang, Paul Pu and others},
  journal={URL https://arxiv. org/abs/2410.23218}
}

@article{qwen3vl,
  title={Qwen3-vl technical report},
  author={Bai, Shuai and Cai, Yuxuan and Chen, Ruizhe and Chen, Keqin and Chen, Xionghui and Cheng, Zesen and Deng, Lianghao and Ding, Wei and Gao, Chang and Ge, Chunjiang and others},
  journal={arXiv preprint arXiv:2511.21631},
  year={2025}
}

@article{scalecua,
  title={Scalecua: Scaling open-source computer use agents with cross-platform data},
  author={Liu, Zhaoyang and Xie, JingJing and Ding, Zichen and Li, Zehao and Yang, Bowen and Wu, Zhenyu and Wang, Xuehui and Sun, Qiushi and Liu, Shi and Wang, Weiyun and others},
  journal={arXiv preprint arXiv:2509.15221},
  year={2025}
}

@article{uitars,
  title={Ui-tars: Pioneering automated gui interaction with native agents},
  author={Qin, Yujia and Ye, Yining and Fang, Junjie and Wang, Haoming and Liang, Shihao and Tian, Shizuo and Zhang, Junda and Li, Jiahao and Li, Yunxin and Huang, Shijue and others},
  journal={arXiv preprint arXiv:2501.12326},
  year={2025}
}

@article{zoomui,
  title={Zoom to Essence: Trainless GUI Grounding by Inferring upon Interface Elements},
  author={Liu, Ziwei and Feng, Tao and Kang, Borui and Yang, Yanbing and Luo, Jun},
  journal={arXiv preprint arXiv:2603.14448},
  year={2026}
}

@article{ui-inst,
  title={UI-Ins: Enhancing GUI Grounding with Multi-Perspective Instruction-as-Reasoning},
  author={Chen, Liangyu and Zhou, Hanzhang and Cai, Chenglin and Zhang, Jianan and Tong, Panrong and Kong, Quyu and Zhang, Xu and Liu, Chen and Liu, Yuqi and Wang, Wenxuan and others},
  journal={arXiv preprint arXiv:2510.20286},
  year={2025}
}

@article{hyperclick,
  title={HyperClick: Advancing Reliable GUI Grounding via Uncertainty Calibration},
  author={Zhang, Shaojie and Fu, Pei and Zhang, Ruoceng and Yang, Jiahui and Du, Anan and Xi, Xiuwen and Wang, Shaokang and Huang, Ying and Qin, Bin and Luo, Zhenbo and others},
  journal={arXiv preprint arXiv:2510.27266},
  year={2025}
}

@article{LPO,
  title={Lpo: Towards accurate gui agent interaction via location preference optimization},
  author={Tang, Jiaqi and Xia, Yu and Wu, Yi-Feng and Hu, Yuwei and Chen, Yuhui and Chen, Qing-Guo and Xu, Xiaogang and Wu, Xiangyu and Lu, Hao and Ma, Yanqing and others},
  journal={arXiv preprint arXiv:2506.09373},
  year={2025}
}

@article{se-gui,
  title={Enhancing visual grounding for gui agents via self-evolutionary reinforcement learning},
  author={Yuan, Xinbin and Zhang, Jian and Li, Kaixin and Cai, Zhuoxuan and Yao, Lujian and Chen, Jie and Wang, Enguang and Hou, Qibin and Chen, Jinwei and Jiang, Peng-Tao and others},
  journal={arXiv preprint arXiv:2505.12370},
  year={2025}
}

@article{tongui,
  title={Tongui: Building generalized gui agents by learning from multimodal web tutorials},
  author={Zhang, Bofei and Shang, Zirui and Gao, Zhi and Zhang, Wang and Xie, Rui and Ma, Xiaojian and Yuan, Tao and Wu, Xinxiao and Zhu, Song-Chun and Li, Qing},
  journal={arXiv e-prints},
  pages={arXiv--2504},
  year={2025}
}

@article{Gui-actor,
  title={Gui-actor: Coordinate-free visual grounding for gui agents},
  author={Wu, Qianhui and Cheng, Kanzhi and Yang, Rui and Zhang, Chaoyun and Yang, Jianwei and Jiang, Huiqiang and Mu, Jian and Peng, Baolin and Qiao, Bo and Tan, Reuben and others},
  journal={arXiv preprint arXiv:2506.03143},
  year={2025}
}

@article{gta1,
  title={Gta1: Gui test-time scaling agent},
  author={Yang, Yan and Li, Dongxu and Dai, Yutong and Yang, Yuhao and Luo, Ziyang and Zhao, Zirui and Hu, Zhiyuan and Huang, Junzhe and Saha, Amrita and Chen, Zeyuan and others},
  journal={arXiv preprint arXiv:2507.05791},
  year={2025}
}

@article{ZwZ,
  title={Zooming without Zooming: Region-to-Image Distillation for Fine-Grained Multimodal Perception},
  author={Wei, Lai and He, Liangbo and Lan, Jun and Dong, Lingzhong and Cai, Yutong and Li, Siyuan and Zhu, Huijia and Wang, Weiqiang and Kong, Linghe and Wang, Yue and others},
  journal={arXiv preprint arXiv:2602.11858},
  year={2026}
}

@article{gupta2026molmoweb,
  title={MolmoWeb: Open Visual Web Agent and Open Data for the Open Web},
  author={Gupta, Tanmay and Wolters, Piper and Ma, Zixian and Sushko, Peter and Pang, Rock Yuren and Llanes, Diego and Yang, Yue and Anderson, Taira and Zheng, Boyuan and Ren, Zhongzheng and others},
  journal={arXiv preprint arXiv:2604.08516},
  year={2026}
}

@article{Propose-then-Critic,
  title={Measure Twice, Click Once: Co-evolving Proposer and Visual Critic via Reinforcement Learning for GUI Grounding},
  author={Wang, Wenkai and Li, Xiyun and Guo, Hongcan and Yu, Wenhao and Fang, Tianqing and Mi, Haitao and Yu, Dong and Zhang, Shengyu},
  journal={arXiv preprint arXiv:2604.21268},
  year={2026}
}

@article{gou2024navigating,
  title={Navigating the digital world as humans do: Universal visual grounding for gui agents},
  author={Gou, Boyu and Wang, Ruohan and Zheng, Boyuan and Xie, Yanan and Chang, Cheng and Shu, Yiheng and Sun, Huan and Su, Yu},
  journal={arXiv preprint arXiv:2410.05243},
  year={2024}
}

@inproceedings{cheng2024seeclick,
  title={Seeclick: Harnessing gui grounding for advanced visual gui agents},
  author={Cheng, Kanzhi and Sun, Qiushi and Chu, Yougang and Xu, Fangzhi and YanTao, Li and Zhang, Jianbing and Wu, Zhiyong},
  booktitle={Proceedings of the 62nd Annual Meeting of the Association for Computational Linguistics (Volume 1: Long Papers)},
  pages={9313--9332},
  year={2024}
}

@article{gui-eyes,
  title={GUI-Eyes: Tool-Augmented Perception for Visual Grounding in GUI Agents},
  author={Chen, Chen and Shao, Jiawei and Lu, Dakuan and Hu, Haoyi and Liu, Xiangcheng and Yao, Hantao and Liu, Wu},
  journal={arXiv preprint arXiv:2601.09770},
  year={2026}
}

@inproceedings{gui-bee,
  title={Gui-bee: Align gui action grounding to novel environments via autonomous exploration},
  author={Fan, Yue and Zhao, Handong and Zhang, Ruiyi and Shen, Yu and Wang, Xin Eric and Wu, Gang},
  booktitle={Proceedings of the 2025 Conference on Empirical Methods in Natural Language Processing},
  pages={33249--33266},
  year={2025}
}

@article{shao2024deepseekmath,
  title={Deepseekmath: Pushing the limits of mathematical reasoning in open language models},
  author={Shao, Zhihong and Wang, Peiyi and Zhu, Qihao and Xu, Runxin and Song, Junxiao and Bi, Xiao and Zhang, Haowei and Zhang, Mingchuan and Li, YK and Wu, Yang and others},
  journal={arXiv preprint arXiv:2402.03300},
  year={2024}
}

@article{guo2025deepseek,
  title={Deepseek-r1: Incentivizing reasoning capability in llms via reinforcement learning},
  author={Guo, Daya and Yang, Dejian and Zhang, Haowei and Song, Junxiao and Wang, Peiyi and Zhu, Qihao and Xu, Runxin and Zhang, Ruoyu and Ma, Shirong and Bi, Xiao and others},
  journal={arXiv preprint arXiv:2501.12948},
  year={2025}
}

@article{yuan2025enhancing,
  title={Enhancing visual grounding for gui agents via self-evolutionary reinforcement learning},
  author={Yuan, Xinbin and Zhang, Jian and Li, Kaixin and Cai, Zhuoxuan and Yao, Lujian and Chen, Jie and Wang, Enguang and Hou, Qibin and Chen, Jinwei and Jiang, Peng-Tao and others},
  journal={arXiv preprint arXiv:2505.12370},
  year={2025}
}

@article{shenfeld2026self,
  title={Self-Distillation Enables Continual Learning},
  author={Shenfeld, Idan and Damani, Mehul and H{\"u}botter, Jonas and Agrawal, Pulkit},
  journal={arXiv preprint arXiv:2601.19897},
  year={2026}
}

@article{qu2026pope,
  title={POPE: Learning to Reason on Hard Problems via Privileged On-Policy Exploration},
  author={Qu, Yuxiao and Setlur, Amrith and Smith, Virginia and Salakhutdinov, Ruslan and Kumar, Aviral},
  journal={arXiv preprint arXiv:2601.18779},
  year={2026}
}

@article{hubotter2026reinforcement,
  title={Reinforcement Learning via Self-Distillation},
  author={H{\"u}botter, Jonas and L{\"u}beck, Frederike and Behric, Lejs and Baumann, Anton and Bagatella, Marco and Marta, Daniel and Hakimi, Ido and Shenfeld, Idan and Buening, Thomas Kleine and Guestrin, Carlos and others},
  journal={arXiv preprint arXiv:2601.20802},
  year={2026}
}

@article{song2026expanding,
  title={Expanding the Capabilities of Reinforcement Learning via Text Feedback},
  author={Song, Yuda and Chen, Lili and Tajwar, Fahim and Munos, Remi and Pathak, Deepak and Bagnell, J Andrew and Singh, Aarti and Zanette, Andrea},
  journal={arXiv preprint arXiv:2602.02482},
  year={2026}
}

@article{song2026survey,
  title={A Survey of On-Policy Distillation for Large Language Models},
  author={Song, Mingyang and Zheng, Mao},
  journal={arXiv preprint arXiv:2604.00626},
  year={2026}
}

@article{zhang2026opsdl,
  title={OPSDL: On-Policy Self-Distillation for Long-Context Language Models},
  author={Zhang, Xinsen and Ding, Zhenkai and Pan, Tianjun and Yang, Run and Kang, Chun and Xiong, Xue and Gu, Jingnan},
  journal={arXiv preprint arXiv:2604.17535},
  year={2026}
}

@article{uivenus15,
  title={UI-Venus-1.5 Technical Report},
  author={Team, Venus and Gao, Changlong and Gu, Zhangxuan and Liu, Yulin and Qiu, Xinyu and Shen, Shuheng and Wen, Yue and Xia, Tianyu and Xu, Zhenyu and Zeng, Zhengwen and others},
  journal={arXiv preprint arXiv:2602.09082},
  year={2026}
}

@article{mobilerl,
  title={Mobilerl: Online agentic reinforcement learning for mobile gui agents},
  author={Xu, Yifan and Liu, Xiao and Liu, Xinghan and Fu, Jiaqi and Zhang, Hanchen and Jing, Bohao and Zhang, Shudan and Wang, Yuting and Zhao, Wenyi and Dong, Yuxiao},
  journal={arXiv preprint arXiv:2509.18119},
  year={2025}
}

@article{computerrl,
  title={Computerrl: Scaling end-to-end online reinforcement learning for computer use agents},
  author={Lai, Hanyu and Liu, Xiao and Zhao, Yanxiao and Xu, Han and Zhang, Hanchen and Jing, Bohao and Ren, Yanyu and Yao, Shuntian and Dong, Yuxiao and Tang, Jie},
  journal={arXiv preprint arXiv:2508.14040},
  year={2025}
}

@article{gui-r1,
  title={Gui-r1: A generalist r1-style vision-language action model for gui agents},
  author={Luo, Run and Wang, Lu and He, Wanwei and Chen, Longze and Li, Jiaming and Xia, Xiaobo},
  journal={arXiv preprint arXiv:2504.10458},
  year={2025}
}

@inproceedings{ui-r1,
  title={Ui-r1: Enhancing efficient action prediction of gui agents by reinforcement learning},
  author={Lu, Zhengxi and Chai, Yuxiang and Guo, Yaxuan and Yin, Xi and Liu, Liang and Wang, Hao and Xiao, Han and Ren, Shuai and Zhao, Pengxiang and Liu, Guangyi and others},
  booktitle={Proceedings of the AAAI Conference on Artificial Intelligence},
  volume={40},
  number={21},
  pages={17608--17616},
  year={2026}
}

@article{li2026rethinking,
  title={Rethinking On-Policy Distillation of Large Language Models: Phenomenology, Mechanism, and Recipe},
  author={Li, Yaxuan and Zuo, Yuxin and He, Bingxiang and Zhang, Jinqian and Xiao, Chaojun and Qian, Cheng and Yu, Tianyu and Gao, Huan-ang and Yang, Wenkai and Liu, Zhiyuan and others},
  journal={arXiv preprint arXiv:2604.13016},
  year={2026}
}

@article{kang2025guirlvg,
  title={GuirlVG: Incentivize GUI Visual Grounding via Empirical Exploration on Reinforcement Learning},
  author={Kang, Weitai and Lei, Bin and Liu, Gaowen and Ding, Caiwen and Yan, Yan},
  journal={arXiv preprint arXiv:2508.04389},
  year={2025}
}

@article{fan2026webfactory,
  title={WebFactory: Automated Compression of Foundational Language Intelligence into Grounded Web Agents},
  author={Fan, Sicheng and Shi, Qingyun and Xu, Shengze and Cai, Shengbo and Zeng, Tieyong and Ling, Li and Shang, Yanyi and Kong, Dehan},
  journal={arXiv preprint arXiv:2603.05044},
  year={2026}
}

@article{huang2025mobileipl,
  title={MobileIPL: Enhancing Mobile Agents Thinking Process via Iterative Preference Learning},
  author={Huang, Kun and Xu, Weikai and Liu, Yuxuan and Wang, Quandong and Gao, Pengzhi and Liu, Wei and Luan, Jian and Wang, Bin and An, Bo},
  journal={arXiv preprint arXiv:2505.12299},
  year={2025}
}

@inproceedings{zeng2026fdc,
  title={FDC-Ground: Improving GRPO for GUI Grounding via Exponential Rewards and Fact-Aligned Pruning},
  author={Zeng, Xiangjian and Li, Wenjing and Wu, Qingqiang and Zhang, Liang},
  booktitle={Proceedings of the AAAI Conference on Artificial Intelligence},
  volume={40},
  number={33},
  pages={28122--28130},
  year={2026}
}

@inproceedings{zhao2026co,
  title={Co-EPG: A Framework for Co-Evolution of Planning and Grounding in Autonomous GUI Agents},
  author={Zhao, Yuan and Zhu, Hualei and Jiang, Tingyu and Li, Shen and Xu, Xiaohang and Wang, Hao Henry},
  booktitle={Proceedings of the AAAI Conference on Artificial Intelligence},
  volume={40},
  number={43},
  pages={36582--36590},
  year={2026}
}

@article{xie2025scaling,
  title={Scaling computer-use grounding via user interface decomposition and synthesis},
  author={Xie, Tianbao and Deng, Jiaqi and Li, Xiaochuan and Yang, Junlin and Wu, Haoyuan and Chen, Jixuan and Hu, Wenjing and Wang, Xinyuan and Xu, Yuhui and Wang, Zekun and others},
  journal={arXiv preprint arXiv:2505.13227},
  year={2025}
}

@article{zhang2025btl,
  title={Btl-ui: Blink-think-link reasoning model for gui agent},
  author={Zhang, Shaojie and Zhang, Ruoceng and Fu, Pei and Wang, Shaokang and Yang, Jiahui and Du, Xin and Cui, Shiqi and Qin, Bin and Huang, Ying and Luo, Zhenbo and others},
  journal={arXiv preprint arXiv:2509.15566},
  year={2025}
}

\clearpage
\appendix
\crefalias{section}{appendix}

\section*{Appendix}
The appendix includes the following aspects:
\begin{itemize}
    \item \Cref{app:benchmarks}: Evaluation Benchmarks.
    \item \Cref{app:training_details}: Training Details.
    \item \Cref{app:additional}: Additional Experiments and Ablations.
\end{itemize}

\section{Evaluation Benchmarks}
\label{app:benchmarks}

\paragraph{ScreenSpot-v2~\cite{atlas}.} 
ScreenSpot-v2 is a refined version of the original ScreenSpot benchmark \cite{cheng2024seeclick}, designed to address annotation ambiguities in earlier versions. It covers mobile, desktop, and web platforms, with each sample consisting of a GUI screenshot paired with a natural language instruction and a ground-truth bounding box. ScreenSpot-v2 is widely adopted as a standard benchmark for general-purpose GUI grounding evaluation.
\paragraph{ScreenSpot-Pro~\cite{screenspotpro}.} 
ScreenSpot-Pro focuses on the under-explored challenge of grounding in professional, high-resolution software environments. It comprises 1,581 instructions captured from 23 real-world applications spanning five professional industries, including development tools (e.g., VSCode, PyCharm), creative applications (e.g., Photoshop, Premiere), CAD/engineering software (e.g., AutoCAD, SolidWorks), scientific tools (e.g., MATLAB, Stata), and office software (e.g., Word, Excel), across three operating systems (Windows, macOS, Linux). The central challenge of ScreenSpot-Pro is the extremely small target size: UI elements occupy on average only 0.07\% of the high-resolution screenshot area. Dense and complex interface layouts further increase the difficulty of precise localization, making ScreenSpot-Pro one of the most demanding grounding benchmarks available.

\paragraph{UI-Vision~\cite{uivision}.} 
UI-Vision is the largest desktop-centric GUI benchmark to date, spanning 83 open-source desktop applications across six domains: Productivity, Development, Creativity, Education, Browsers, and Entertainment. We evaluate on its Element Grounding benchmark, which contains over 8,200 query-label pairs with high-quality human-annotated bounding boxes. A distinguishing aspect of UI-Vision is its cross-application diversity, requiring GUI agents to generalize across highly varied software interfaces and interaction patterns, exposing limitations in spatial reasoning and professional software understanding.

\paragraph{OSWorld-G and OSWorld-G-Refine~\cite{osworldg}.} 
OSWorld-G is a comprehensive GUI grounding benchmark set in the Linux environment, comprising 564 finely annotated samples that cover 32 distinct UI element types. OSWorld-G captures diverse real-world computer-use interactions, requiring software knowledge, layout understanding, and fine-grained operations. The benchmark organizes tasks into four categories: text matching, element recognition, layout understanding, and precise operation. OSWorld-G-Refine is a refined version that rewrites instructions to remove domain-specific knowledge, isolating the model's pure spatial grounding ability from its software understanding.

\paragraph{MMBench GUI L2~\cite{mmbench}.} 
MMBench-GUI is a hierarchical, cross-platform benchmark for evaluating GUI automation agents across six platforms: Windows, macOS, Linux, iOS, Android, and Web. The benchmark is organized into four ascending capability levels: L1-Content Understanding, L2-Element Grounding, L3-Task Automation, and L4-Task Collaboration. In our evaluation, we adopt the L2-Element Grounding level, which assesses the model's ability to localize target GUI elements from natural language instructions. L2 includes both basic and advanced difficulty tiers with diverse instruction styles (e.g., action descriptions, target element descriptions, and refusal cases). Its unique contribution lies in enabling consistent cross-platform comparison under a unified evaluation protocol, revealing how models handle the varying interface designs and visual layouts across different operating systems.

\section{Training Details.}
\label{app:training_details}
\paragraph{Training Data.} 
Our training data is sourced entirely from the grounding training subset of ScaleCUA~\cite{scalecua}. Since the original annotations contain labeling errors and lack instruction diversity, we apply a two-stage data curation pipeline. First, we leverage UI-Venus1.5-8B~\cite{uivenus15} to filter the dataset, retaining only samples where its prediction agrees with the original annotation, removing noisy labels. Second, we employ Qwen3-VL-8B~\cite{qwen3vl} to rewrite the original instructions into diverse paraphrases, enriching instruction variety. After filtering and rewriting, approximately 7K samples remain for training.

\paragraph{Hyperparameters.} 
The hyper-parameter details for GUI-SD are provided in \Cref{tab:hyper}
\begin{table}[t]
\centering
\caption{Hyperparameter settings for training.}
\label{tab:hyperparameters}
\begin{tabular}{lc}
\toprule
\textbf{Hyperparameter} & \textbf{Value} \\
\midrule
Learning Rate & from 2.5e-6 to 0 \\
Per Device Train Batch Size & 1 \\
Gradient Accumulation Steps & 16 \\
Number of Training Epochs & 1 \\
Warmup Ratio & 0.05 \\
Maximum Sequence Length & 20000 \\
Maximum Completion Length & 128 \\
EMA Decay Coefficient & 0.95 \\
DeepSpeed (Student) & ZeRO-2 \\
DeepSpeed (Teacher) & ZeRO-3 \\
$\sigma$ Scale Factor & 1.5 \\
Minimum $\sigma$ Floor & $\sqrt{0.1} \cdot \min(W, H)$ \\

\bottomrule
\end{tabular}
\label{tab:hyper}
\end{table}

\section{Additional Experiments and Ablations}
\label{app:additional}
\subsection{Detailed Benchmark Results}
\label{app:detailed_results}
We provide per-category results for each evaluation benchmark. Table~\ref{tab:ssp} reports ScreenSpot-Pro results across six professional domains (CAD, Development, Creative, Scientific, Office, OS), split by Text and Icon targets. Table~\ref{tab:ss2} reports ScreenSpot-v2 results across Mobile, Desktop, and Web platforms. Table~\ref{tab:uivision} reports UI-Vision results across Basic, Functional, and Spatial grounding tasks. Table~\ref{tab:mmbench} reports MMBench GUI L2 results across six operating systems (Windows, macOS, Linux, iOS, Android, Web), split by Basic and Advanced difficulty. Table~\ref{tab:osworldg} and Table~\ref{tab:osworldgr} report OSWorld-G and OSWorld-G-Refine results across four task types (Text Matching, Element Recognition, Layout Understanding, Fine-Grained Manipulation). GUI-SD consistently outperforms all GRPO baselines across the majority of sub-categories.

\begin{table*}[t]
\centering
\caption{Performance comparison on the ScreenSpot-Pro benchmark \cite{screenspotpro}.}
\vspace{5pt}
\resizebox{\textwidth}{!}{%
\begin{tabular}{l cc cc cc cc cc cc c}
\toprule
\multirow{2}{*}{\textbf{Methods}} & \multicolumn{2}{c}{\textbf{CAD}} & \multicolumn{2}{c}{\textbf{Development}} & \multicolumn{2}{c}{\textbf{Creative}} & \multicolumn{2}{c}{\textbf{Scientific}} & \multicolumn{2}{c}{\textbf{Office}} & \multicolumn{2}{c}{\textbf{OS}} & \multirow{2}{*}{\textbf{Avg.}} \\
\cmidrule(lr){2-3} \cmidrule(lr){4-5} \cmidrule(lr){6-7} \cmidrule(lr){8-9} \cmidrule(lr){10-11} \cmidrule(lr){12-13}
 & \textbf{Text} & \textbf{Icon} & \textbf{Text} & \textbf{Icon} & \textbf{Text} & \textbf{Icon} & \textbf{Text} & \textbf{Icon} & \textbf{Text} & \textbf{Icon} & \textbf{Text} & \textbf{Icon} & \\
\midrule
\textit{Qwen3-VL-Instruct} \cite{qwen3vl} & 58.38 & 14.06 & 79.22 & 24.83 & 69.70 & 17.48 & 76.39 & 27.27 & 80.79 & 35.85 & 71.03 & 26.97 & 53.57 \\
\textit{+ GRPO-Binary} \cite{infigui-r1} & 63.45 & 20.31 & 80.52 & 31.72 & 70.71 & 18.88 & 79.17 & 31.82 & 84.18 & 37.74 & 72.90 & 30.34 & 56.80 \\
\textit{+ GRPO-Distance} \cite{se-gui} & 62.44 & 18.75 & 82.47 & 29.66 & 70.71 & 19.58 & 78.47 & 33.64 & 83.62 & 39.62 & 71.03 & 30.34 & 56.61 \\
\textit{+ GRPO-Gaussian} \cite{GUI-G$^2$} & 62.44 & 21.88 & 84.42 & 30.34 & 71.21 & 20.28 & 79.86 & 34.55 & 84.18 & 39.62 & 71.96 & 29.21 & 57.37 \\
\midrule
\textbf{+ Ours} & \textbf{67.01} & \textbf{31.25} & \textbf{83.77} & \textbf{35.86} & \textbf{73.74} & \textbf{22.38} & \textbf{82.64} & \textbf{37.27} & \textbf{84.75} & \textbf{52.83} & \textbf{72.90} & \textbf{37.08} & \textbf{60.72} \\
\bottomrule
\end{tabular}
}%
\label{tab:ssp}
\end{table*}

\begin{table*}[t]
\centering
\caption{Performance comparison on the Screenspotv2 benchmark \cite{atlas}.}
\resizebox{0.75\textwidth}{!}{%
\begin{tabular}{l cc cc cc c}
\toprule
\multirow{2}{*}{\textbf{Methods}} & \multicolumn{2}{c}{\textbf{Mobile}} & \multicolumn{2}{c}{\textbf{Desktop}} & \multicolumn{2}{c}{\textbf{Web}} & \multirow{2}{*}{\textbf{Avg.}} \\
\cmidrule(lr){2-3} \cmidrule(lr){4-5} \cmidrule(lr){6-7}
 & \textbf{Text} & \textbf{Icon} & \textbf{Text} & \textbf{Icon} & \textbf{Text} & \textbf{Icon} & \\
\midrule
\textit{Qwen3-VL-Instruct} \cite{qwen3vl} & 97.93 & 88.63 & 96.91 & 89.29 & 95.30 & 87.68 & 93.16 \\
\textit{+ GRPO-Binary} \cite{infigui-r1} & 99.66 & 89.57 & 98.97 & 89.29 & 96.15 & 90.15 & 94.58 \\
\textit{+ GRPO-Distance } \cite{se-gui} & 99.66 & 88.63 & 97.94 & 89.29 & 96.15 & 86.70 & 93.71 \\
\textit{+ GRPO-Gaussian} \cite{GUI-G$^2$} & 99.66 & 89.10 & 97.42 & 89.29 & 95.73 & 88.67 & 93.95 \\
\midrule
\textbf{+ Ours} & \textbf{99.66} & \textbf{89.57} & \textbf{97.42} & \textbf{92.86} & \textbf{97.01} & \textbf{91.13} & \textbf{95.05} \\
\bottomrule
\end{tabular}
}%
\label{tab:ss2}
\end{table*}

\begin{table*}[!htbp]
\centering
\caption{Performance comparison on the UI-Vision benchmark \cite{uivision}.}
\vspace{5pt}
\resizebox{0.55\textwidth}{!}{%
\begin{tabular}{l cccc}
\toprule
\textbf{Methods} & \textbf{Basic} & \textbf{Functional} & \textbf{Spatial} & \textbf{Avg.} \\
\midrule
\textit{Qwen3-VL-Instruct} \cite{qwen3vl} & 30.53 & 31.88 & 14.16 & 25.19 \\
\textit{+ GRPO-Binary} \cite{infigui-r1} & 34.03 & 34.59 & 15.35 & 27.61 \\
\textit{+ GRPO-Distance} \cite{se-gui}& 34.09 & 33.97 & 15.45 & 27.47 \\
\textit{+ GRPO-Gaussian} \cite{GUI-G$^2$}  & 35.16 & 34.76 & 15.76 & 28.18 \\
\midrule
\textbf{+ Ours} & \textbf{41.87} & \textbf{39.39} & \textbf{19.79} & \textbf{33.27} \\
\bottomrule
\end{tabular}
}%
\label{tab:uivision}
\end{table*}

\begin{table*}[!htbp]
\centering
\caption{Performance comparison on the MMBench-GUI L2 benchmark \cite{mmbench}.}
\vspace{5pt}
\resizebox{\textwidth}{!}{%
\begin{tabular}{l cc cc cc cc cc cc c}
\toprule
\multirow{2}{*}{\textbf{Methods}} & \multicolumn{2}{c}{\textbf{Windows}} & \multicolumn{2}{c}{\textbf{MacOS}} & \multicolumn{2}{c}{\textbf{Linux}} & \multicolumn{2}{c}{\textbf{iOS}} & \multicolumn{2}{c}{\textbf{Android}} & \multicolumn{2}{c}{\textbf{Web}} & \multirow{2}{*}{\textbf{Avg.}} \\
\cmidrule(lr){2-3} \cmidrule(lr){4-5} \cmidrule(lr){6-7} \cmidrule(lr){8-9} \cmidrule(lr){10-11} \cmidrule(lr){12-13}
 & \textbf{Bas.} & \textbf{Adv.} & \textbf{Bas.} & \textbf{Adv.} & \textbf{Bas.} & \textbf{Adv.} & \textbf{Bas.} & \textbf{Adv.} & \textbf{Bas.} & \textbf{Adv.} & \textbf{Bas.} & \textbf{Adv.} & \\
\midrule
\textit{Qwen3-VL-Instruct} \cite{qwen3vl} & 89.30 & 65.07 & 85.51 & 70.81 & 76.96 & 58.16 & 95.86 & 84.24 & 96.35 & 85.63 & 95.48 & 77.60 & 82.99 \\
\textit{+ GRPO-Binary} \cite{infigui-r1} & 90.04 & 67.65 & 86.67 & 71.97 & 76.44 & 61.73 & 96.18 & 86.36 & 96.35 & 87.89 & 95.48 & 80.52 & 84.32 \\
\textit{+ GRPO-Distance} \cite{se-gui}& 91.88 & 69.12 & 83.19 & 69.36 & 75.39 & 58.16 & 96.18 & 84.85 & 96.07 & 87.61 & 94.52 & 78.25 & 83.27 \\
\textit{+ GRPO-Gaussian} \cite{GUI-G$^2$}  & 91.88 & 68.38 & 83.19 & 69.65 & 75.39 & 58.67 & 96.50 & 86.06 & 96.07 & 88.17 & 95.16 & 80.52 & 83.71 \\
\midrule
\textbf{+ Ours} & \textbf{91.14} & \textbf{72.06} & \textbf{90.72} & \textbf{76.30} & \textbf{78.01} & \textbf{66.33} & \textbf{97.13} & \textbf{86.67} & \textbf{96.91} & \textbf{91.27} & \textbf{95.81} & \textbf{83.44} & \textbf{86.65} \\
\bottomrule
\end{tabular}%
}
\label{tab:mmbench} 
\end{table*}

\begin{table*}[t]
\centering
\caption{Performance comparison on the OSWorld-G benchmark \cite{osworldg}.}
\vspace{5pt}
\resizebox{0.75\textwidth}{!}{%
\begin{tabular}{l ccccc}
\toprule
\textbf{Methods} & \makecell{\textbf{Text}\\\textbf{Matching}} & \makecell{\textbf{Element}\\\textbf{Recognition}} & \makecell{\textbf{Layout}\\\textbf{Understanding}} & \makecell{\textbf{Fine-Grained}\\\textbf{Manipulation}} & \textbf{Avg.} \\
\midrule
\textit{Qwen3-VL-Instruct} \cite{qwen3vl} & 47.37 & 67.16 & 66.67 & 62.12 & 58.69 \\
\textit{+ GRPO-Binary} \cite{infigui-r1} & 47.37 & 70.90 & 70.22 & 62.88 & 61.17 \\
\textit{+ GRPO-Distance} \cite{se-gui} & 42.11 & 74.63 & 70.67 & 62.88 & 62.06 \\
\textit{+ GRPO-Gaussian} \cite{GUI-G$^2$} & 42.11 & 73.13 & 70.67 & 63.64 & 61.88 \\
\midrule
\textbf{+ Ours} & \textbf{52.63} & \textbf{75.37} & \textbf{73.78} & \textbf{63.64} & \textbf{64.01} \\
\bottomrule
\end{tabular}
}%
\label{tab:osworldg}
\end{table*}

\begin{table*}[t]
\centering
\caption{Performance comparison on the OSWorld-G-Refine benchmark \cite{osworldg}.}
\vspace{5pt}
\resizebox{0.75\textwidth}{!}{%
\begin{tabular}{l ccccc}
\toprule
\textbf{Methods} & \makecell{\textbf{Text}\\\textbf{Matching}} & \makecell{\textbf{Element}\\\textbf{Recognition}} & \makecell{\textbf{Layout}\\\textbf{Understanding}} & \makecell{\textbf{Fine-Grained}\\\textbf{Manipulation}} & \textbf{Avg.} \\
\midrule
\textit{Qwen3-VL-Instruct} \cite{qwen3vl} & 52.63 & 78.36 & 79.56 & 65.15 & 67.38 \\
\textit{+ GRPO-Binary} \cite{infigui-r1} & 47.37 & 79.85 & 79.11 & 70.45 & 68.62 \\
\textit{+ GRPO-Distance} \cite{se-gui} & 47.37 & 82.09 & 80.00 & 71.97 & 69.86 \\
\textit{+ GRPO-Gaussian} \cite{GUI-G$^2$}  & 52.63 & 82.09 & 80.00 & 71.97 & 70.04 \\
\midrule
\textbf{+ Ours} & \textbf{52.63} & \textbf{82.09} & \textbf{83.56} & \textbf{69.70} & \textbf{70.92} \\
\bottomrule
\end{tabular}
}%
\label{tab:osworldgr}
\end{table*}

\subsection{Ablation on Visual Privilege Design}
Beyond the main ablation in \Cref{tab:table3}, we further compare different visual privilege designs in Table~\ref{tab:visual_privilege}. The first row is the Naive OPSD baseline that appends the target coordinate as text. The second row replaces textual privilege with a standard zoom that masks all regions outside a fixed-size area centered on the target while drawing a bounding box on it. The third row uses our adaptive zoom with a hard mask that adjusts the visible region proportionally to the ground-truth bounding box size. Both visual privilege variants substantially outperform Naive OPSD (+4.5 and +4.7), confirming that delivering ground-truth information through the visual channel is critical. The adaptive zoom achieves a further gain over standard zoom, as its flexible masking better adapts to varying target sizes across different GUI layouts.

\begin{table}[t]
\centering
\caption{Ablation on visual privilege design. SSP denotes ScreenSpot-Pro accuracy. $\Delta$ OPSD denotes the performance difference relative to Naive OPSD.}
\label{tab:visual_privilege}

\begin{tabular}{ll|cc}
\toprule
\multicolumn{2}{c|}{\textbf{Teacher's Guidance Setting}} & \multicolumn{2}{c}{\textbf{Student}} \\
\textbf{Visual Context} & \textbf{Text Context} & \textbf{SSP} & \textbf{$\Delta$ OPSD} \\
\midrule
Orig. Image & Inst. + Text BBox & 55.6 & 0 \\
Standard Zoom + Drawn BBox & Inst. + Drawn Hint & 60.1 & +4.5 \\
Adaptive Zoom + Drawn BBox & Inst. + Drawn Hint & 60.3 & +4.7 \\
\bottomrule
\end{tabular}

\end{table}

\subsection{The Self-teacher Improves during Training}
A key design choice in GUI-SD is how the teacher model is maintained during training. Unlike off-policy distillation where the teacher is typically frozen, on policy self-distillation allows the teacher to evolve alongside the student. As shown in Table~\ref{tab:teacher_update}, using the current policy $q_{\theta}$ directly as the teacher yields 59.4\%. In this setting, the teacher updates simultaneously with the student at every step, resulting in minimal divergence between their distributions and weakening the distillation signal. Freezing the teacher at initialization ($q_{\theta_{\text{ref}}}$) performs similarly at 59.6\%, as the fixed teacher quickly becomes outdated once the student improves beyond its initial capacity. Exponential moving average (EMA) with a decay of 0.95 achieves the best result (60.7\%), balancing stability and adaptability by allowing the teacher to gradually absorb the student's improving policy while maintaining a smoother, more reliable distribution. A lower decay of 0.90 updates the teacher more slowly, causing it to lag behind the student's progress, reducing performance to 59.8\%.

\begin{table}[h]
\centering
\caption{Ablation on teacher update strategy. $q_{\theta}$ denotes using the current policy as the teacher (updated every step), $q
_{\theta_{\text{ref}}}$ denotes using a frozen copy of the initial policy as the teacher, and EMA denotes the exponential moving average teacher with the specified decay coefficient.}
\label{tab:teacher_update}
\begin{tabular}{lc}
\toprule
Teacher & ScreenSpot-Pro \\
\midrule
$q_{\theta}$ & 59.4 \\
$q_{\theta_{\text{ref}}}$ & 59.6 \\
EMA = 0.90 & 59.8 \\
EMA = 0.95 & \textbf{60.7} \\
\bottomrule
\end{tabular}
\end{table}

\subsection{Performance Across Model Sizes}
As shown in Table~\ref{tab:scaling}, GUI-SD consistently improves over the base model across all three scales of Qwen3-VL-Instruct (2B, 4B, and 8B). On ScreenSpot-Pro, GUI-SD achieves gains of +3.7, +4.6, and +7.1 at the 2B, 4B, and 8B scales respectively. Similar improvements are observed across all other benchmarks, confirming that the proposed method is effective regardless of model capacity.

\begin{table}[]
\centering
\caption{Performance comparison across six representative grounding benchmarks for Qwen3-VL from 2B to 8B. ScreenSpot-Pro (SSP), ScreenSpot-v2 (SS2), UI-Vision (UIV), OSWorld-G (OSW-G), OSWorld-G-Refine (OSW-GR), and MMBench GUI L2 (MMG).}
\label{tab:scaling}
\resizebox{0.8\textwidth}{!}{
\begin{tabular}{lcccccc}
\toprule
\textbf{Method} & \textbf{SSP} & \textbf{SS2} & \textbf{UIV} & \textbf{OSW-G} & \textbf{OSW-GR} & \textbf{MMG} \\
\midrule
Qwen3-VL-Instruct-2B \cite{qwen3vl} & 39.4 & 86.4 & 13.3 & 46.6 & 62.2 & 70.3 \\
+ \textit{Ours} & \textbf{43.1} & \textbf{90.7} & \textbf{19.4} & \textbf{49.8} & \textbf{62.6} & \textbf{75.2} \\
\midrule
Qwen3-VL-Instruct-4B \cite{qwen3vl} & 56.1 & 92.7 & 23.8 & 62.8 & 70.9 & 83.1 \\
+ \textit{Ours} & \textbf{60.7} & \textbf{94.0} & \textbf{31.5} & \textbf{63.3} & \textbf{71.1} & \textbf{85.2} \\
\midrule
Qwen3-VL-Instruct-8B \cite{qwen3vl} & 53.6 & 93.2 & 25.2 & 58.7 & 67.4 & 83.0 \\
+ \textit{Ours} & \textbf{60.7} & \textbf{95.1} & \textbf{33.3} & \textbf{64.0} & \textbf{70.9} & \textbf{86.7} \\
\bottomrule
\end{tabular}
}
\end{table}



\end{document}